\documentclass[letterpaper, 10 pt, journal, twoside]{IEEEtran}  

\IEEEoverridecommandlockouts                              

\usepackage{algorithm}
\usepackage{algpseudocodex}  
\usepackage{amsmath}
\usepackage{amssymb}
\usepackage{booktabs}
\usepackage{cite}
\usepackage{graphicx}
\usepackage[hidelinks]{hyperref}  
\usepackage{leftindex}
\usepackage{mathtools}  
\usepackage[lofdepth,lotdepth,caption=false]{subfig}  
\usepackage{siunitx}
\usepackage{array}
\usepackage{xcolor}
\usepackage{makecell}
\usepackage{flushend}

\usepackage{pdfpages}  

\renewcommand{\arraystretch}{1.2} 
\makeatletter
\newcommand\fs@betterruled{%
  \def\@fs@cfont{\bfseries}\let\@fs@capt\floatc@ruled
  \def\@fs@pre{\vspace*{7pt}\hrule height.8pt depth0pt \kern2pt}%
  \def\@fs@post{\kern2pt\hrule\relax}%
  \def\@fs@mid{\kern2pt\hrule\kern2pt}%
  \let\@fs@iftopcapt\iftrue}
\floatstyle{betterruled}
\restylefloat{algorithm}
\makeatother

\usepackage{amsmath,amsfonts,bm}

\def\Secref#1{Section~\ref{#1}}

\def\1{\bm{1}}

\def\vdelta{{\bm{\delta}}}
\def\vzero{{\bm{0}}}

\def\vtheta{{\bm{\theta}}}

\def\vh{{\bm{h}}}

\def\vk{{\bm{k}}}

\def\vp{{\bm{p}}}

\def\vx{{\bm{x}}}

\def\mR{{\bm{R}}}

\def\mT{{\bm{T}}}

\DeclareMathAlphabet{\mathsfit}{\encodingdefault}{\sfdefault}{m}{sl}
\SetMathAlphabet{\mathsfit}{bold}{\encodingdefault}{\sfdefault}{bx}{n}

\def\sQ{{\mathbb{Q}}}

\newcommand{\R}{\mathbb{R}}

\def\frame{\Sigma}
\def\jointspace{\sQ}
\def\fk#1{\textnormal{FK} \left( #1 \right)}
\def\ik#1{\textnormal{IK} \left( #1 \right)}

\def\specialOrthogonal{\textnormal{SO}\left( 3 \right) }
\def\transform{\mT}
\def\vtheta{\boldsymbol{\theta}}

\def\desired{\transform_D}

\newcommand{\axjnt}[1]{\vh_{#1}}
\newcommand{\intersect}[2]{\vk_{#1}}
\newcommand{\pjnt}[2]{ \leftindex^{#1} \vp_{#2}}
\newcommand{\rotmat}[2]{ \leftindex^{#1} \mR_{#2}}
\newcommand{\rotmatconst}[2]{ \leftindex^{#1} \mR_{#2}^*}

\newcommand{\xhdrNoPeriod}[1]{\vspace{1.7mm}\noindent{{\bf #1}}}

\usepackage{letltxmacro}
\LetLtxMacro{\originaleqref}{\eqref}
\renewcommand{\eqref}[1]{\originaleqref{#1}}
\newcommand{\Eqref}[1]{\originaleqref{#1}}

\newcommand{\figref}[1]{Fig.~\ref{#1}}
\newcommand{\tabref}[1]{Table~\ref{#1}}

\newcommand{\daniel}[1]{#1}

\title{\LARGE \bf
Automatic Geometric Decomposition for\\Analytical Inverse Kinematics
}

\ifdefined \anonymous
\author{Anonymous authors according to the IEEE RAS Guidelines for double-anonymous submissions}
\else
\author{
Daniel Ostermeier,
Jonathan K\"ulz,
and Matthias Althoff
\thanks{Received 7 May 2025; accepted 18 July 2025, 2025; Date of publication 11 August 2025.
This article was recommended for publication by Associate Editor L. Scalera and Editor L. Pallottino upon evaluation of the Associate Editor and Reviewers' comments.
This work was supported by the Deutsche Forschungsgemeinschaft (German Research Foundation) under grant number AL 1185/31-1.
\textit{(Corresponding Author: Daniel Ostermeier)}
}
\thanks{Daniel Ostermeier is with the Department of Computer Engineer
ing, Technical University of Munich, 85748 Garching, Germany (e-mail:
 daniel.sebastian.ostermeier@tum.de).}%
\thanks{Jonathan K\"ulz and Matthias Althoff are with the Department of Computer
 Engineering, Technical University of Munich, 85748 Garching, Germany, and
 also with the Munich Center for Machine Learning (MCML), 80333 Munich,
 Germany (e-mail: jonathan.kuelz@tum.de; althoff@tum.de).}
\thanks{Digital Object Identifier (DOI): 10.1109/LRA.2025.3597897}
}
\fi

\begin{document}
\bstctlcite{IEEEexample:BSTcontrol}  

\maketitle
\IEEEpubid{
  \parbox{\textwidth}{\centering\footnotesize
    \vspace{10pt}\textcopyright~2025 The Authors. This work is licensed under a Creative Commons Attribution 4.0 License.
    For more information, see \url{https://creativecommons.org/licenses/by/4.0/}%
  }%
}
\IEEEpubidadjcol

\begin{abstract}
    Calculating the inverse kinematics (IK) is a fundamental challenge in robotics.
    Compared to numerical or learning-based approaches, analytical IK provides higher efficiency and accuracy.
    However, existing analytical approaches are difficult to use in most applications, as they require human ingenuity in the derivation process, are numerically unstable, or rely on time-consuming symbolic manipulation.
    In contrast, we propose a method that, for the first time, enables an analytical IK derivation and computation in less than a millisecond in total.
    Our work is based on an automatic online decomposition of the IK into pre-solved, numerically stable subproblems via a kinematic classification of the respective manipulator.
    In numerical experiments, we demonstrate that our approach is orders of magnitude faster in deriving the IK than existing tools that employ symbolic manipulation.
    Following this one-time derivation, our method matches and often surpasses baselines, such as IKFast, in terms of speed and accuracy during the computation of explicit IK solutions.
    Finally, we provide an open-source C++ toolbox with Python wrappers that substantially reduces the entry barrier to using analytical IK in applications like rapid prototyping and kinematic robot design.
\end{abstract}

\section{Introduction}
\IEEEPARstart{I}{nverse} kinematics (IK) describes the problem of finding joint angles of a kinematic chain for a given end-effector pose.
The IK problem is known to be analytically solvable for a specific set of kinematic chains that fulfill certain geometric properties~\cite{Pieper}.
Industrial robots are often designed with these properties in mind.
In these cases, human-driven manual derivation of an analytical IK is time-consuming but feasible, as it is only done once during the entire engineering process.
Recently, computational design methods for monolithic~\cite{Sorokin2023, Vaish2024} or modular robots~\cite{Whitman2020, Kuelz2024} have proven promising in optimizing industrial robots. 
They are based on iterative optimization and feature many kinematically distinct manipulators in dynamic problem settings.
Manually deriving an analytical IK for each possible manipulator design is time-consuming, requires domain expertise, and, hence, is usually infeasible.
Existing methods that automatically derive the analytical IK either rely on symbolic manipulation---which takes several minutes per manipulator---or are numerically unstable in the vicinity of singularities and workspace boundaries.
Due to these limitations of existing analytical solvers, less accurate, computationally intensive numerical methods that yield incomplete solutions are a typical choice in the mentioned applications, even if analytical solutions are known to exist.
\begin{figure}[!t]
    \vspace{0.06cm}
    \centering
    \includegraphics[width=0.9\linewidth]{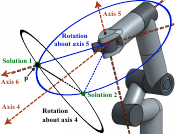}
    \caption{\daniel{When solving the IK for a 6-DOF robot with a spherical wrist, we obtain the angles for its axes four and five via the visualized subproblem. A unit circle is hereby centered around axes four (black) and five (blue). Each unit circle is contained in a plane whose normal vector aligns with the respective axis.
    Both planes contain the point $p$: A unit offset originating from the intersection point of the last three axes (red dot) along axis six.
    The intersection points of the two unit circles resemble the potential solutions to this subproblem.
    The solutions to all other angles are obtained in a similar fashion. 
        Invalid solutions, i.e., Solution 2, are discarded in the process.}}
    \label{fig:SP2Visualized}
\end{figure}
\IEEEpubidadjcol
Our work represents the first approach to enable analytical IK in real-time computational manipulator design, e.g., for modular robots.
The proposed method is based on geometric feature matching against a set of known kinematic classes.
Through this classification process, we rapidly identify analytically solvable kinematic chains.
The IK problem of these chains is decomposed into a series of geometric subproblems.
Both the kinematic classification (conducted once per manipulator) and the subsequent evaluation of the subproblems (conducted once per desired end-effector pose) are real-time-capable and do not require human intervention or symbolic manipulation.
We further propose an automated kinematic remodeling procedure that makes our method applicable to arbitrary parameterizations of the kinematic chain, such as a Unified Robot Description Format~(URDF) file, Denavit-Hartenberg~(DH) parameters \cite{DH-Params}, or homogeneous coordinates of the joint frames in the zero configuration.
Our method is currently limited to non-redundant manipulators with revolute joints.
For redundant manipulators (e.g., $\geq$7R), our method can be used to search for a set of joints that, when locked in place, allow us to analytically solve the resulting kinematics.
To solve the subproblems that we obtain from our decomposition algorithm, we leverage the numerically stable solutions introduced by Elias et al.~\cite{elias_canonical}, which we extend to an even larger share of kinematic classes than initially considered.
\daniel{We implement our method in our open-source toolbox Efficient Analytical Inverse Kinematics (EAIK).
Additional visualizations and an introduction guide are available on our project website:}
\ifdefined \anonymous
\mbox{\url{https://pub-eaik.github.io/pub-eaik}}.
\else 
\mbox{\url{https://eaik.cps.cit.tum.de}}.
\fi
\section{Related Work}\label{sec:relatedWork}
\begin{table}
\centering
\caption{Comparison of predominant methods for analytical IK.}
    \begin{tabular}{@{}l@{\hspace{4mm}}c@{\hspace{4mm}}c@{\hspace{4mm}}c@{\hspace{4mm}}c@{}}
    \toprule
    & Ours & IKFast \cite{rosenDiankov} & \thead{Eigenstructure \\ \cite{genericIKEigen, efficientEigendecomposition}} & \thead{Elias et al. \\ \cite{elias_canonical}} \\
    \midrule
    Automatic derivation & Yes & Yes & Yes & No\\
    Numerically stable & Yes & Yes & $^*$ & Yes\\
    Code available & Yes & Yes & No & Yes\\
    Classes \figref{fig:KinematicClasses} & 3, 5--9 & 2--9$^{**}$  & 2--9$^{**}$ & 8,9 \\
    \bottomrule
    \multicolumn{5}{l}{\footnotesize $^*$Numerical stability is doubted\daniel{~\cite{rosenDiankov, IKBT}}.}\\
    \multicolumn{5}{l}{\footnotesize $^{**}$The actual limits of these methods are insufficiently known.}
    \end{tabular}\\
\label{tab:RelatedWorkMethodComparison}
\vspace{-2mm}
\end{table}

Most approaches for solving the IK problem can be split into analytical, numerical, and learning-based methods.
Numerical methods involve iterative algorithms that gradually refine an initial guess for the joint configuration until a desired end-effector pose is achieved.
To this end, many approaches use the inverse or pseudo-inverse of the manipulator Jacobian~\cite[Chapter 6.2]{LynchPark}.
Alternatively, the IK can be described as a high-dimensional minimization problem, so arbitrary numerical optimization algorithms can be deployed, optionally, under consideration of additional constraints~\cite{fabrik}.
Compared to analytical approaches, numerical IK has an increased computation time, can suffer from numerical instability, and usually only provides a single solution~\cite{dktpart1}.
Learning-based methods, such as IKFlow~\cite{Ames2022}, can be advantageous for redundant manipulators~\cite{Bensadoun2022} by drawing diverse samples from infinite solution spaces.
However, these methods do not generalize well across different manipulators, which renders them useless if the manipulator topology is not known a priori.

While a general solution for analytical IK is yet to be discovered, a wide range of works~\cite{Ho2012, jointLockArmAngle,Xiao2011} focuses on hand-derived IK formulations for specific manipulators.
Fundamental aspects of a more broadly applicable method date back to the 1960s, when Donald L. Pieper \cite{Pieper} proposed a polynomial-based approach alongside two sufficient conditions---nowadays known as ``Pieper criteria''---for the existence of an analytical solution:
Assuming a non-degenerate robot, he states that any 6R manipulator is solvable given that either
(a) the last or first three joint axes of a manipulator intersect in a common point (as shown in \figref{fig:SP2Visualized}) or
(b) three consecutive intermediate joint axes of the robot are parallel.
Furthermore, all non-degenerate 3R manipulators are known to be analytically solvable~\cite{Reconfigurable}.
The only widely acknowledged necessary condition for the existence of an analytical IK is for the manipulator chain to be non-degenerate~\cite{IKBT}. 
Degeneracy occurs if the last link of a kinematic chain has fewer degrees of freedom than the kinematic chain has joints---either for specific end-effector poses (singularities) or for all possible poses within its workspace (implicit redundancies)~\cite{Pieper}.
While manipulators with seven or more degrees of freedom are inherently degenerate, solutions can be derived by joint-locking \cite{redundantModular, lockedJointFailure, redundantWorkspace} or a fixed (arm) angle in the case of elbow manipulators~\cite{elias7DOF, 7dofIK, fixedArmAngle, jointLockArmAngle}.
For non-degenerate manipulators,~\cite{conformalGeometricAlgebra} proposes an approach based on conformal geometric algebra to solve the IK for robots with spherical wrists.
A similar framework is proposed in~\cite{nRRobotsGeometricAlgebra}, where a semi-analytical method for manipulators with arbitrary rotational joints is developed.

A common approach for deriving IK solutions is decomposing the problem into geometric subproblems~\cite{Paden, Kahan}.
Subproblems define a distinct set of geometric correspondences for which the analytical solution is known a priori.
One seeks to re-obtain these subproblems as decoupled, individually solvable terms in the IK formulation of a manipulator by exploiting collinearities and intersections between certain joint axes.
Previous work proposes different subproblem representations~\cite{studyComparisonSubproblems,dualQuaternions}. 
Further canonical sets of pre-solved subproblems are presented in~\cite{Paden, Kahan, elias_canonical, extension2ndSubproblem}.
In this work, we build directly on the canonical set proposed by Elias et al.~\cite{elias_canonical}.
While they derive the IK formulations for certain 6R manipulators, their derivations depend on coinciding reference points for intersecting joint axes, which often require tedious manual reformulations of the manipulator kinematics.
Furthermore, the authors do not elaborate on how a feasible decomposition can be automatically chosen for a given manipulator.
\begin{figure}
    \vspace{0.06cm}
    \centering
    \includegraphics[width=\linewidth]{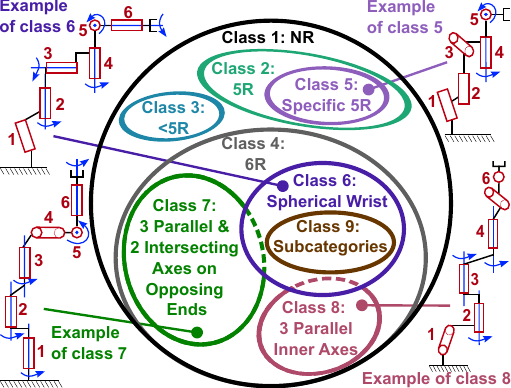}
    \caption{We assign classes to certain sets of kinematic chains. 
    All depicted classes except 1, 2, and 4 are solvable by our method.
    The subproblem decompositions for classes 8 and 9 have already been derived in~\cite{elias_canonical}.
    Further distinctions within the classes are necessary for the decompositions in the \hyperref[sec:newDecompositions]{Appendix}, but are not depicted for simplicity.}
    \label{fig:KinematicClasses}
\end{figure}

Approaches on general solvers, such as IKFast~\cite{rosenDiankov}, aim to automate the analytical IK derivations.
IKFast relies on symbolic manipulation of the IK equations and tangent-half-angle substitution to derive respective univariate polynomials as proposed by Raghavan and Roth~\cite{RaghavenRoth} and, e.g., adopted in~\cite{Manocha92}. 
In \cite{comparisonRaghavanSubproblems}, the approach of Raghavan and Roth is compared to a subproblem-based IK formulation, \daniel{where the latter showed higher accuracy, robustness, and computational efficiency.}
The IK problem may also be defined as an eigenstructure problem, as shown by \cite{genericIKEigen, efficientEigendecomposition}.
The numerical stability of these eigenstructure-based approaches, however,  is doubted in more recent publications \cite{rosenDiankov, IKBT}.
Another take on general symbolic solvers is presented in~\cite{IKBT}, where behavior trees are used to break up the symbolic IK formulations following a rule-based approach.
In~\cite{SolutionExistence}, the authors 
propose an algorithm similar to the subproblem decompositions presented in our work, but with the sole motivation of deciding on analytical solvability.

Prior work on manipulator design focuses on kinematic properties such as static reachability~\cite{Campos2019, Althoff2019, Whitman2020, Romiti2023, Kuelz2023, Kuelz2024}, dexterity~\cite{Leibrandt2023}, or coverage of a pre-defined workspace~\cite{Liu2020, Hoffman2025}.
Efficiently evaluating these properties requires solving the IK for many novel manipulator designs.
While analytical solvers are theoretically applicable in this context, all existing general solvers exhibit one of two limitations in this context: either the derivation process, e.g., generating a C++ file, takes multiple minutes, or the resulting solver is considered unstable in the vicinity of workspace boundaries or singularities.
Our work addresses these limitations by combining fast derivations with stable solutions for a broad class of manipulators.

\section{Notation and Problem Statement}\label{sec:prerequisites}
\subsection{Notation}\label{sec:Notation}
This work is limited to kinematic chains of rigid bodies connected by revolute joints.
To represent the kinematics, we employ an exponential coordinate representation for rotations~\cite[Chapter 3.2.3]{LynchPark}, paired with three-dimensional Cartesian displacement vectors that represent the position offset between two joint axes.

\daniel{Let $\mathbb{S}^1$ represent the unit circle in the plane and for $n \in \mathbb{N}_{+}$: $\jointspace=\mathbb{S}_1^1\times\dots\times\mathbb{S}^1_{n-1}$~\cite[Chapter 3.2.1]{mathRobotic}.}
The kinematics of a manipulator with $n$ revolute joints and joint space \daniel{$\jointspace$} are fully defined by $n$ joint axes with respect to a basis frame $\frame_0$, and the static orientation of the end effector relative to the last joint axis. 
Each joint axis is specified by a unit vector \mbox{$\axjnt{j} \in \R^3$} that is expressed in the basis frame, and a reference point \mbox{$\pjnt{0}{j} \in \R^3$} on the axis.
We write $\pjnt{j}{k}$ to denote the translational offset between the reference points of axes $j$ and $k$.
$\mR(\vh_j,\theta_j) \in \specialOrthogonal$ denotes the rotation matrix  that corresponds to a rotation about the unit vector $\vh_j$ by angle $\theta_j$.
We omit the argument for readability and write \mbox{$\rotmat{j-1}{j}$} for the remainder of this paper to denote the rotation induced by rotating the $j$-th joint.
We generalize this notation to $\rotmat{i}{k}$, where $i < k - 1$, to represent consecutive rotations about $\axjnt{i+1}, \dots, \axjnt{k}$ by the corresponding joint angles.
Further, \mbox{$\rotmat{0}{EE} = \rotmat{0}{j} \rotmat{j}{EE}$} represents the rotation between the base frame and the end-effector frame given the static rotation $\rotmat{j}{EE}$.

\subsection{Problem statement}
The forward kinematics (FK) function \mbox{$\textnormal{FK}: \jointspace \rightarrow SE(3)$} of the manipulator returns the end-effector pose \mbox{$\mT_{EE} \in SE(3)$} for the joint angles $\vtheta\in\jointspace$ and can be computed in various ways, e.g., using the product of exponentials formula or the equation of Rodrigues~\cite[Chapter 4]{LynchPark}.
We define the position FK according to~\eqref{eqn:position_fk} and the orientation FK via~\eqref{eqn:orientation_fk}.
\begin{align}
    \pjnt{0}{EE} &= \pjnt{0}{1} + \left(  \sum_{i=1}^{n-1} \rotmat{0}{i} \pjnt{i}{i+1} \right) + \rotmat{0}{n} \pjnt{n}{EE} \label{eqn:position_fk}
\\
    \rotmat{0}{EE} &= \left(  \prod_{i=0}^{n-1}\rotmat{i}{i+1} \right) \rotmat{n}{EE} \label{eqn:orientation_fk}
\end{align}
Given a desired pose $\desired$, we want to find the discrete set of \underline{all} analytically feasible solutions to the IK problem
\begin{align}
    \ik{\desired} = \left\{
        \vtheta \in \jointspace \mid \fk{\vtheta} = \desired
    \right\} \, .
\end{align}
We do not impose any order on the obtained solutions.


\section{Method}
\label{sec:SubproblemDecomposition}
\begin{figure}
    \vspace{0.06cm}
    \centering
    \includegraphics[scale=1]{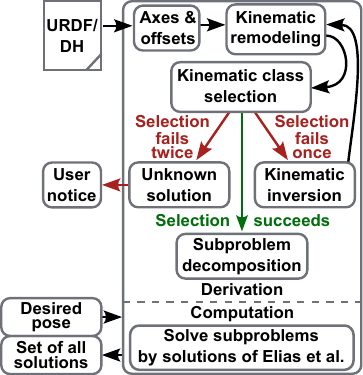}%
    \caption{
    Flow diagram: Assigning a manipulator to a kinematic class.
    }
    \label{fig:MethodOverview}
\end{figure}

We assume that the kinematic structure of a manipulator follows the convention in \Secref{sec:Notation}, which can be easily obtained from a URDF\footnote{Elaboration on the URDF: \url{https://wiki.ros.org/urdf}} file, DH parameters \cite{DH-Params}, or likewise.
Prior works~\cite{elias_canonical, studyComparisonSubproblems, dualQuaternions} showed that the IK problem for specific manipulators is solvable by decomposing its kinematic formulation into analytically solvable subproblems by hand.
To automate this process, we start by \textit{remodeling} the given kinematic chain.
We thereby shift the reference points on the joint axes such that the following properties realize a simplification of $\ik{\desired}$:
\begin{itemize}
    \item According to \Eqref{eqn:position_fk}, the position kinematics of a manipulator is invariant to a rotation \mbox{${}^0\mR_j$} whenever $\pjnt{j}{j+1} = \vzero$.
    \item If two or more joint axes are parallel or anti-parallel, their unit vectors $\axjnt{i}$ can be chosen equally by negating the rotation angle, and consecutive rotations can be grouped in a single rotation according to the rotation formula of Rodrigues~\cite[Chapter 4]{LynchPark}.
\end{itemize}
After remodeling, we assign each manipulator to a kinematic class (see \figref{fig:KinematicClasses}).

An overview of our proposed method is shown in~\figref{fig:MethodOverview}.
Subproblem decompositions for classes 8 and 9 are given by~\cite{elias_canonical}. 
Additionally, we obtain decompositions for class 3 and classes 5--7 using the derivations presented in the \hyperref[sec:newDecompositions]{Appendix} and on our website. 
If a subproblem decomposition is not known for the resulting class, we invert the chain as described in \Secref{sec:KinematicInversion} and try again.
If the remodeling and inversion procedure yields a class with a known decomposition, we can obtain the complete analytical solution set for arbitrary end-effector poses in the workspace. 
We refer to remodeling, inversion, and class assignment as the derivation process.

To cope with analytically unsolvable manipulators (e.g., with seven or more joints), we can employ an~\textit{uninformed search} for a set of joints that, when locked in place, yields a solvable manipulator.
This is done by iterating over all possible combinations of locked joints and applying the derivation algorithm in~\figref{fig:MethodOverview}.
Up until now, to the best of our knowledge, such a search algorithm could not be employed with mathematical certainty (i.e., via analytical expressions) in a reasonable time and is only made feasible by the fast derivation times of our method. 

\subsection{Kinematic remodeling}
\begin{figure}
    \vspace{0.06cm}
    \centering
    \includegraphics[width=\linewidth]{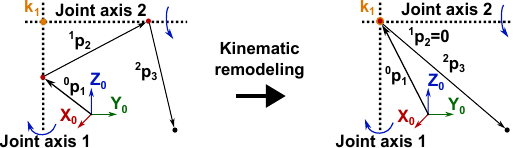}
    \caption{Kinematic remodeling for the example of two intersecting axes: We translate the reference points (red dots) along the corresponding axes until they coincide. Black dots mark reference points of the subsequent axes.}
    \label{fig:kinematicRemodeling}
\end{figure}
The detection of parallel and  anti-parallel axes is straightforward.
Translational offsets between the reference points on the joint axes, on the other hand, are typically non-zero for common parametrizations of real-world manipulators.
We obtain a zero-valued offset for two or more consecutive intersecting axes by moving their corresponding reference points to the point where the axes intersect, as visualized in \figref{fig:kinematicRemodeling}.
We compute the intersection point of two axes by following the approach from~\cite{graphicsGems}.
If two axes do not intersect perfectly, e.g., due to numerical inaccuracies, it yields the closest point to both axes.

We denote the intersection point of two joint axes $j$ and $j+1$ as $\intersect{j}{j+1} \in \mathbb{R}^3$. 
As $\intersect{j}{j+1}$ is located on both joint axes $j$ and $j+1$, we can shift their reference points $\pjnt{0}{j}$ and $\pjnt{0}{j+1}$ to coincide with $\intersect{j}{j+1}$.
As a result, \mbox{$\pjnt{j}{j+1} = \vzero$}.
Both $\rotmat{j-1}{j}$ and $\rotmat{j}{j+1}$ are hereby left unchanged, and so is \Eqref{eqn:orientation_fk}.
However, we need to further assure that \Eqref{eqn:position_fk} is invariant to our remodeling.
We hence adjust the preceding and subsequent displacement vectors:
\begin{align}
    {}^{j-1}\vp_{j}     &= \vk_j-{}^{0}\vp_{j-1} \label{eq:RemodelDisplacement}\\ 
    {}^{j}\vp_{j+1}     &= \vzero \\ 
    {}^{j+1}\vp_{j+2}   &= {}^{0}\vp_{j+2}-\vk_j
    \,.
\end{align}
If the first two joint axes intersect ($j=0$), we can omit the adjustment \daniel{in} \Eqref{eq:RemodelDisplacement} by choosing our base frame such that its origin coincides with the first reference point.
This procedure is not always unique, as an axis can intersect with both the preceding and succeeding axes in the kinematic chain, but at different intersection points.
In this case, we choose the first representation to which we can apply a known decomposition. 

\subsection{Subproblem decompositions}
\label{sec:Subproblems}
Singular configurations of the robot, or end-effector poses in the proximity of workspace boundaries, often pose an issue to analytical methods due to numerical instabilities.
We employ the subproblems introduced in~\cite{elias_canonical}, which are posed such that approximate least-squares\footnote{These least-squares formulations minimize the error of each subproblem. The solutions may not correspond to the global least-squares formulation for the Euclidean distance between the end effector and desired pose.} solutions are obtained if no finite set of exact solutions exists. Continuous and stable IK solutions are thereby guaranteed even when facing singularities.
\daniel{The authors in~\cite{elias_canonical} further derive subproblem decompositions for predominant 6R manipulator families, and hint at some additional cases for manipulators containing specific intersecting or parallel axes.
We adapt their derivations in our method and complement their work by providing explicit solutions to the missing cases in the \hyperref[sec:newDecompositions]{Appendix}.}

Non-degenerate 1R, 2R, 3R, and 4R manipulators (class 3) are solvable even without additional intersecting or parallel axes.
On the other hand, 5R manipulators are solvable per our method if contained in class 5, i.e., if they meet at least one of the following sufficient criteria:
\begin{itemize}
    \item The last or first two axes 1,2 (5,6) intersect.
    \item The intermediate axes 2,3 (3,4) intersect while the axes 3,4 (2,3) are parallel.
    \item Any three consecutive axes are parallel.
\end{itemize}
The derivations for all manipulators with fewer than six joints follow the same scheme we applied in the \hyperref[sec:newDecompositions]{Appendix}.
For brevity, we refrain from explicitly denoting the obtained decompositions and their derivations here, but instead present them on our project website in a separate document:
\ifdefined \anonymous
\mbox{\url{https://pub-eaik.github.io/pub-eaik}}.
\else 
\mbox{\url{https://eaik.cps.cit.tum.de}}.
\fi

\subsection{Kinematic inversion}\label{sec:KinematicInversion}
We extend the introduced derivations to an even broader set of manipulators, i.e., those that are created by switching end effector and basis of the derived manipulators, by inverting their kinematic chain.
Hereby, we redefine:
\begin{align}
    \axjnt{j}' = -\axjnt{n-j+1}, &\quad j \in \{1, \dots, n\} \\
    \pjnt{j}{j+1}' = -\pjnt{n-j}{n-j+1}, &\quad j \in \{1, \dots, n\}
    \,.
\end{align}
Hence, we obtain new position and orientation forward kinematics for the redefined joint axes with
\begin{align}
    \pjnt{0}{n}' &= \sum_{k=n}^{1} \rotmat{k}{k-1} \pjnt{k}{k-1} =-\rotmat{0}{n}^T \pjnt{0}{n}
    \label{eq:InvertedforwardPositionKin} \\
    \rotmat{0}{n}' &= \prod_{k=n}^{1} \rotmat{k}{k-1} = \rotmat{0}{n}^T
    \label{eq:InvertedforwardOrientationKin}
    \,.
\end{align}
The joint angles are invariant with respect to the inversion procedure.
Hence, we can compute any IK problem of the original (non-inverted) kinematic chain by solving the IK of its inverted chain for the inverse of the original end-effector pose.


\section{Experimental Evaluation and Comparison}\label{sec:evaluation}
We compare our method to different frameworks on common 6R manipulators:
the UR5 robot (three parallel and two intersecting axes, \underline{class 8}), the Puma (spherical wrist and two intersecting axes, \underline{class 9}), and the ABB IRB 6640 (spherical wrist and two parallel axes, \underline{class 9}).
A Franka Emika Panda (spherical wrist after kinematic inversion, \underline{class 6}) represents the category of redundant 7R robots, which we solve by locking joint~7.
We further use two fictitious manipulators from~\cite{elias_canonical}: The ``Spherical" (spherical wrist, \underline{class 9}) and ``3-Parallel" (three parallel axes, \underline{class 8}) robot.
Although derivations for these manipulators were already shown in~\cite{elias_canonical}, the URDF and DH parameterizations used in our experiments did not naturally exhibit coinciding joint frames at axis intersection points.
Hence, the original derivations fail when applied directly and rely on a tedious reformulation of the manipulator kinematics by a human.
Our method resolves this issue via kinematic remodeling and inversion. 
We compare the computation speed and accuracy of our method to that of IKFast~\cite{rosenDiankov}, the learning-based IKFlow method \cite{Ames2022}, as well as the numerical Gauss-Newton~(GN) and Levenberg-Marquardt~(LM) solvers implemented in the Python Robotics Toolbox~\cite{rtb}. 
Our experiments with IKBT~\cite{IKBT} showed certain problems ranging from slow file generation to non-functional code. 
In our opinion, its strength lies in the fully symbolic output in the form of LaTeX code.
Other publicly available solvers, like YAIK\footnote{YAIK implementation: \url{https://github.com/weigao95/yaik}}, show promising results but lack an underlying scientific publication and basic documentation. As IKFast is well-maintained and the current standard way of deriving analytical IK, it serves as the baseline against which we compare.
All experiments were conducted on an AMD~Ryzen~7 8-core CPU.

\subsection{Computational efficiency}
\label{sec:computationalEfficiency}
\tabref{tab:derivationTimes} shows the initial derivation time to obtain an analytical IK solution for our approach and IKFast.
For our method, we also distinguish between an optimized DH parametrization with direct access to the underlying C++ implementation and a URDF-based parametrization that includes larger parts of unoptimized Python code to obtain the representation we define in \Secref{sec:Notation}.
The latter allows for a fairer comparison to IKFast, which depends on a similar file format.
In the case of IKFast, the derivation time includes the duration of code generation without compilation.
For our method, the measured time includes a single run of the forward kinematics calculation to obtain the positions of all joint axes in zero-pose, the remodeling of the kinematic chain, and the kinematic class selection.
The results clearly show the main advantage of our method (EAIK) over existing tools:
The automatic remodeling and decomposition into subproblems is orders of magnitude faster than the automatic code generation of IKFast.
In the specific example of the UR5 robot, IKFast takes more than 15 minutes compared to \SI{39}{\micro\second} using EAIK.
The mean derivation time of our method was below \SI{50}{\micro\second} for all evaluated 6R robots.

\begin{figure*}
    \vspace{0.06cm}
    \centering
    \subfloat[]{%
        \includegraphics{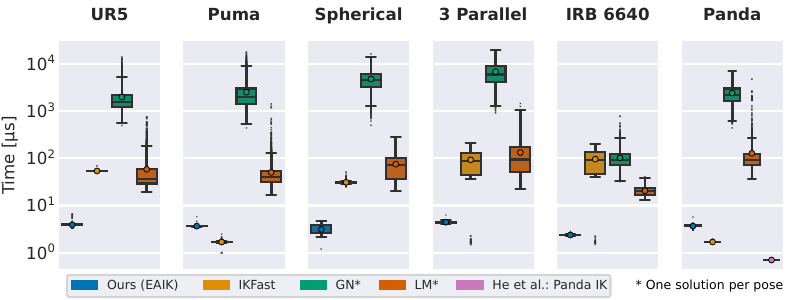}%
        \label{fig:speed_comparison_a}%
    }
    \hfill
    \subfloat[]{%
        \includegraphics{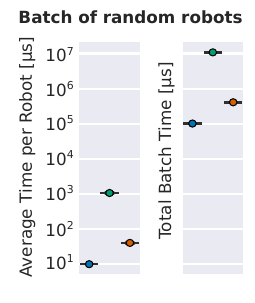}%
        \label{fig:speed_comparison_b}%
    }
    \caption{Comparison of the IK computation times of five representative manipulators on 5,000 randomly assigned end-effector poses (a) and the batch-times on 10,000 random (analytically solvable) 6R manipulators (b). The measured times in (a) correspond to a single solution for the numerical approaches---\daniel{which we marked with an asterisk}---and the set of all possible solutions for EAIK, IKFast, and the method by He et al.~\cite{PandaSpecificIK}, respectively.
    In (b), we include derivation times and only show our method along with the numerical ones, as the derivation times of IKFast are too long for this task to finish within a reasonable time.
    Our method shows the smallest overall variance and, except for the Puma and Panda robot, consistently surpasses all other methods.}
    \label{fig:speed_comparison}
\end{figure*}

After derivation, we generate 5,000 IK problems for each robot by calculating the forward kinematics for randomly sampled joint angles.
\figref{fig:speed_comparison_a} shows the computation times for IKFast, both numerical approaches, and EAIK.
We also compare the methods with respect to the 7R Franka Panda manipulator and include timings for its hand-derived analytical IK by He et al.~\cite{PandaSpecificIK}.
All of the analytical methods rely on joint locking for the Panda, so we set the last joint to a fixed random value.
Except for the Puma, the Panda, and some rare outliers, where IKFast performs extraordinarily well, our implementation manages to surpass the computation speed of IKFast by a factor of at least five.
The measured times correspond to a single solution for the numerical approaches and the set of all possible solutions for EAIK, IKFast, and He et al., respectively.
As expected, both numerical solvers are significantly slower than EAIK and IKFast.
The highly specialized method by He et al. serves as a baseline for how fast manual implementations are when tailored to a specific manipulator. 
It performs best for the Panda but inherently fails when applied to one of the other presented manipulators.

Additionally, we evaluate a scenario inspired by modular robot design, where IK solutions are required for a large variety of different manipulators~\cite{Whitman2020, Kuelz2024, Althoff2019}.
We generate a batch of 10,000 6R manipulators (belonging to analytically solvable kinematic classes) via randomized DH parameters and sample one random pose from the workspace of each manipulator.
Prior to our work, long derivation times disqualified most analytical methods from being used in such challenging tasks, with numerical methods remaining the only option---even if analytically solvable robots, as in~\cite{Althoff2019}, make up the major part of the design space.
The combined time for IK derivation and computation is displayed in~\figref{fig:speed_comparison_b}, which shows that our method is consistently faster than both numerical methods, all while computing the full solution set.

\begin{table}
\centering
\caption{Derivation times for analytical IK solutions.}
    \begin{tabular}{@{}lccccccc@{}}
    \toprule
     & Ours (URDF) & Ours (DH) & IKFast \\
    \midrule
    IRB6640 & 36\textmu s & \textbf{3.7\textmu s} & 22$\cdot$10\textsuperscript{6}\textmu s \\
    Spherical & 36\textmu s & \textbf{3.9\textmu s} & 16$\cdot$10\textsuperscript{6}\textmu s \\
    UR5 & 39\textmu s & \textbf{3.6\textmu s} & 95$\cdot$10\textsuperscript{7}\textmu s \\
    Puma & 35\textmu s & \textbf{3.8\textmu s} & 37$\cdot$10\textsuperscript{6}\textmu s \\
    3-Parallel & 36\textmu s & \textbf{3.3\textmu s} & 53$\cdot$10\textsuperscript{7}\textmu s \\
    Panda (Joint 7 locked) & 45\textmu s & \textbf{6.3\textmu s} & 93$\cdot$10\textsuperscript{6}\textmu s \\
    \bottomrule
    \end{tabular}
\label{tab:derivationTimes}
\vspace{-2mm}
\end{table}
To compare our approach to the learning-based IKFlow, we again choose the \daniel{7R} Panda robot.
While comparison on a \daniel{6R} robot would be more representative, the authors do not provide pre-trained networks for such manipulators.
Our own endeavors in training their network to the UR5 resulted in position errors in the order of multiple centimeters, which we deem not representative.
To solve the Panda analytically, we lock joint seven in a random position.
IKFlow is able to make use of all seven joints.
Our experiments show a mean computation time of \SI{75.07}{\micro\second} for IKFlow (\SI{7507}{\micro\second} per batch of 100 poses) and \SI{3.67}{\micro\second} for our method.
IKFlow is only able to perform batched computations.  

\subsection{Accuracy}
The accuracy of analytical methods (i.e., ours and IKFast) is only limited by the chosen floating-point representation and the numeric errors that originate from mathematical operations within it.
The solutions we obtain via IKFlow for the Panda manipulator result in a median position error of \SI{7.3e-3}{\meter}.
The median position error of our approach is \SI{1.12e-15}{\meter} using the same poses as with IKFlow on the Panda.
The error of our method thereby lies within the expected double-precision floating-point accuracy.
When evaluated on the UR5 robot, the numerical methods (GN and LM) show a similar median position error of about $10^{-4}$ m.
This is multiple orders of magnitude higher than the median error of our method ($10^{-15}$ m) and IKFast ($10^{-12}$ m) for the same experimental setup.
The error on the numerical solvers vastly depends on the number of iterations undergone (up to 30 in our case) and poses a trade-off between accuracy and computation time.
A quantitative comparison of different numerical solvers and their respective parameters is available in \cite{dktpart1, dktpart2}.
\section{Conclusion}
\label{Conclusion}
Our approach for solving IK problems based on subproblem decomposition is automatic, real-time capable, and numerically stable.
Current analytical solvers either require manual derivations at some point, rely on slow symbolic manipulation, or are numerically unstable.
Our open-source implementation substantially lowers the entry barrier for using analytical IK.
It facilitates the use of analytical IK in domains where the properties of a kinematic chain change frequently, such as in (modular) robot design, where derivation times of multiple seconds or even minutes rendered analytical approaches unfeasible in the past.

While we present an extensive set of analytically solvable kinematic classes, there may exist more sufficient conditions that allow us to decompose novel kinematic classes to obtain analytical solutions.
Our method is currently incapable of solving parallel kinematic chains, manipulators with joints that are not revolute, or robotic systems with multiple end effectors.
Further, joint limits, trajectory consistency, and collision constraints need to be verified a posteriori by computing the full-body forward kinematics for the identified solutions.
Possible future work includes: \daniel{(a) the integration of (semi-)numerical fallbacks for manipulators with no known subproblem decompositions, e.g., via the search-based approach in~\cite{elias_canonical}}, \daniel{(b)} explicit parametrizations for detected internal redundancies, and \daniel{(c)} solutions for manipulators with linear axes or parallel joint configurations.



\bibliographystyle{IEEEtran}
\bibliography{IEEEabrv, refs_clean, ieeetransettings} 

\begin{thebibliography}{10}
\providecommand{\url}[1]{#1}
\csname url@rmstyle\endcsname
\providecommand{\newblock}{\relax}
\providecommand{\bibinfo}[2]{#2}
\providecommand\BIBentrySTDinterwordspacing{\spaceskip=0pt\relax}
\providecommand\BIBentryALTinterwordstretchfactor{4}
\providecommand\BIBentryALTinterwordspacing{\spaceskip=\fontdimen2\font plus
\BIBentryALTinterwordstretchfactor\fontdimen3\font minus
  \fontdimen4\font\relax}
\providecommand\BIBforeignlanguage[2]{{%
\expandafter\ifx\csname l@#1\endcsname\relax
\typeout{** WARNING: IEEEtran.bst: No hyphenation pattern has been}%
\typeout{** loaded for the language `#1'. Using the pattern for}%
\typeout{** the default language instead.}%
\else
\language=\csname l@#1\endcsname
\fi
#2}}

\bibitem{Pieper}
D.~L. Pieper, ``The kinematics of manipulators under computer control,'' Ph.D.
  dissertation, Department of Mechanical Engineering, Stanford University,
  1968.

\bibitem{Sorokin2023}
M.~Sorokin, \emph{et~al.}, ``On designing a learning robot: Improving
  morphology for enhanced task performance and learning,'' in \emph{Proc. of
  the IEEE/RSJ Int. Conf. on Intelligent Robots and Systems (IROS)}, 2023, pp.
  487--494.

\bibitem{Vaish2024}
A.~Vaish and O.~Brock, ``Co-designing manipulation systems using task-relevant
  constraints,'' in \emph{Proc. of the IEEE Int. Conf. on Robotics and
  Automation (ICRA)}, 2024, pp. 4177--4183.

\bibitem{Whitman2020}
J.~Whitman, \emph{et~al.}, ``Modular robot design synthesis with deep
  reinforcement learning,'' in \emph{Proc. of the AAAI Conf. on Artificial
  Intelligence (AAAI)}, vol.~34, no.~06, 2020, pp. 10\,418--10\,425.

\bibitem{Kuelz2024}
J.~K\"{u}lz and M.~Althoff, ``Optimizing modular robot composition: A
  lexicographic genetic algorithm approach,'' in \emph{Proc. of the IEEE Int.
  Conf. on Robotics and Automation (ICRA)}, 2024, pp. 16\,752--16\,758.

\bibitem{DH-Params}
J.~Denavit and R.~S. Hartenberg, ``A kinematic notation for lower-pair
  mechanisms based on matrices,'' \emph{Applied Mechanics}, vol.~22, no.~2, pp.
  215--221, 1955.

\bibitem{elias_canonical}
A.~J. Elias and J.~T. Wen, ``{IK-Geo}: Unified robot inverse kinematics using
  subproblem decomposition,'' \emph{Mechanism and Machine Theory}, vol. 209,
  no. 105971, 2025.

\bibitem{rosenDiankov}
R.~Diankov, ``Automated construction of robotic manipulation programs,'' Ph.D.
  dissertation, Robotics Institute, Carnegie Mellon University, 2010.

\bibitem{genericIKEigen}
D.~Manocha and Y.~Zhu, ``A fast algorithm and system for the inverse kinematics
  of general serial manipulators,'' in \emph{Proc. of the IEEE Int. Conf. on
  Robotics and Automation (ICRA)}, vol.~4, 1994, pp. 3348--3353.

\bibitem{efficientEigendecomposition}
D.~Manocha and J.~Canny, ``Efficient inverse kinematics for general {6R}
  manipulators,'' \emph{Transactions on Robotics and Automation}, vol.~10,
  no.~5, pp. 648--657, 1994.

\bibitem{IKBT}
D.~Zhang and B.~Hannaford, ``{IKBT}: Solving symbolic inverse kinematics with
  behavior tree,'' \emph{Artificial Intelligence Research}, vol.~65, pp.
  457--486, 2019.

\bibitem{LynchPark}
K.~M. Lynch and F.~C. Park, \emph{Modern Robotics: Mechanics, Planning, and
  Control}, 1st~ed.\hskip 1em plus 0.5em minus 0.4em\relax Cambridge University
  Press, 2017.

\bibitem{fabrik}
A.~Aristidou and J.~Lasenby, ``Fabrik: A fast, iterative solver for the inverse
  kinematics problem,'' \emph{Graphical Models}, vol.~73, no.~5, pp. 243--260,
  2011.

\bibitem{dktpart1}
J.~Haviland and P.~Corke, ``Manipulator differential kinematics: Part 1:
  Kinematics, velocity, and applications,'' \emph{Robotics \& Automation
  Magazine}, pp. 2--11, 2023.

\bibitem{Ames2022}
B.~Ames, J.~Morgan, and G.~Konidaris, ``{IKFlow}: Generating diverse inverse
  kinematics solutions,'' \emph{IEEE Robotics and Automation Letters}, vol.~7,
  no.~3, 2022.

\bibitem{Bensadoun2022}
R.~Bensadoun, \emph{et~al.}, ``Neural inverse kinematics,'' in \emph{Proc. of
  the Int. Conf. on Machine Learning (ICML)}, vol. 162, 2022, pp. 1787--1797.

\bibitem{Ho2012}
T.~Ho, C.-G. Kang, and S.~Lee, ``Efficient closed-form solution of inverse
  kinematics for a specific six-{DOF} arm,'' \emph{Control, Automation and
  Systems}, vol.~10, no.~3, pp. 567--573, 2012.

\bibitem{jointLockArmAngle}
B.~Tondu, ``A closed-form inverse kinematic modelling of a 7{R} anthropomorphic
  upper limb based on a joint parametrization,'' in \emph{Proc. of the IEEE-RAS
  Int. Conf. on Humanoid Robots (Humanoids)}, 2006, pp. 390--397.

\bibitem{Xiao2011}
W.~Xiao, \emph{et~al.}, ``Closed-form inverse kinematics of 6{R} milling robot
  with singularity avoidance,'' \emph{Production Engineering}, vol.~5, no.~1,
  pp. 103--110, 2011.

\bibitem{Reconfigurable}
I.-M. Chen and Y.~Gao, ``Closed-form inverse kinematics solver for
  reconfigurable robots,'' in \emph{Proc. of the IEEE Int. Conf. on Robotics
  and Automation (ICRA)}, 2001, pp. 2395--2400.

\bibitem{redundantModular}
R.~C. Luo, T.-W. Lin, and Y.-H. Tsai, ``Analytical inverse kinematic solution
  for modularized 7-{DOF} redundant manipulators with offsets at shoulder and
  wrist,'' in \emph{Proc. of the IEEE/RSJ Int. Conf. on Intelligent Robots and
  Systems (IROS)}, 2014, pp. 516--521.

\bibitem{lockedJointFailure}
W.~Xu, Y.~She, and Y.~Xu, ``Analytical and semi-analytical inverse kinematics
  of {SSRMS}-type manipulators with single joint locked failure,'' \emph{Acta
  Astronautica}, vol. 105, no.~1, pp. 201--217, 2014.

\bibitem{redundantWorkspace}
I.~Zaplana and L.~Basanez, ``A novel closed-form solution for the inverse
  kinematics of redundant manipulators through workspace analysis,''
  \emph{Mechanism and Machine Theory}, vol. 121, pp. 829--843, 2018.

\bibitem{elias7DOF}
A.~J. Elias and J.~T. Wen, ``Redundancy parameterization and inverse kinematics
  of 7-{DOF} revolute manipulators,'' \emph{Mechanism and Machine Theory}, vol.
  204, no. 105824, 2024.

\bibitem{7dofIK}
M.~Shimizu, \emph{et~al.}, ``Analytical inverse kinematic computation for
  7-{DOF} redundant manipulators with joint limits and its application to
  redundancy resolution,'' \emph{Transactions on Robotics}, vol.~24, no.~5, pp.
  1131--1142, 2008.

\bibitem{fixedArmAngle}
H.~Moradi and S.~Lee, ``Joint limit analysis and elbow movement minimization
  for redundant manipulators using closed form method,'' in \emph{Proc. of the
  Int. Conf. on Intelligent Computing (ICIC)}, 2005, pp. 423--432.

\bibitem{conformalGeometricAlgebra}
I.~Zaplana, H.~Hadfield, and J.~Lasenby, ``Closed-form solutions for the
  inverse kinematics of serial robots using conformal geometric algebra,''
  \emph{Mechanism and Machine Theory}, vol. 173, no. 104835, 2022.

\bibitem{nRRobotsGeometricAlgebra}
Y.~Wei, \emph{et~al.}, ``General approach for inverse kinematics of {nR}
  robots,'' \emph{Mechanism and Machine Theory}, vol.~75, pp. 97--106, 2014.

\bibitem{Paden}
B.~E. Paden, ``Kinematics and control of robot manipulators,'' Ph.D.
  dissertation, EECS Department, University of California, Berkeley, 1986.

\bibitem{Kahan}
W.~Kahan, ``Lecture on computational aspects of geometry,'' University of
  California, Berkeley, 1983.

\bibitem{studyComparisonSubproblems}
E.~Sariyildiz, E.~Cakiray, and H.~Temeltas, ``A comparative study of three
  inverse kinematic methods of serial industrial robot manipulators in the
  screw theory framework,'' \emph{Advanced Robotic Systems}, vol.~8, no.~5, pp.
  9--24, 2011.

\bibitem{dualQuaternions}
P.-F. Lin, M.-B. Huang, and H.-P. Huang, ``Analytical solution for inverse
  kinematics using dual quaternions,'' \emph{IEEE Access}, vol.~7, pp.
  166\,190--166\,202, 2019.

\bibitem{extension2ndSubproblem}
T.~Yue-sheng and X.~Ai-ping, ``Extension of the second {Paden-Kahan}
  sub-problem and its' application in the inverse kinematics of a
  manipulator,'' in \emph{Proc. of the IEEE Conf. on Robotics, Automation and
  Mechatronics (RAM)}, 2008, pp. 379--381.

\bibitem{RaghavenRoth}
M.~Raghaven and B.~Roth, ``Kinematic analysis of the 6{R} manipulator of
  general geometry,'' in \emph{Proc. of the Int. Symp. on Robotics Research
  (ISRR)}, 1991, pp. 263--269.

\bibitem{Manocha92}
D.~Manocha and J.~Canny, ``Real time inverse kinematics for general {6R}
  manipulators,'' in \emph{Proc. of the IEEE Int. Conf. on Robotics and
  Automation (ICRA)}, vol.~1, 1992, pp. 383--389.

\bibitem{comparisonRaghavanSubproblems}
T.~Truong, \emph{et~al.}, ``General solution for inverse kinematics of six
  degrees of freedom of a welding robot arm,'' in \emph{Proc. of the Int. Conf.
  on System Science and Engineering (ICSSE)}, 2023, pp. 387--394.

\bibitem{SolutionExistence}
W.~Shanda, \emph{et~al.}, ``Existence conditions and general solutions of
  closed-form inverse kinematics for revolute serial robots,'' \emph{Applied
  Sciences}, vol.~9, no.~20, p. 4365, 2019.

\bibitem{Campos2019}
T.~Campos, \emph{et~al.}, ``Task-based design of ad-hoc modular manipulators,''
  in \emph{Proc. of the IEEE Int. Conf. on Robotics and Automation (ICRA)},
  2019, pp. 6058--6064.

\bibitem{Althoff2019}
M.~Althoff, \emph{et~al.}, ``Effortless creation of safe robots from modules
  through self-programming and self-verification,'' \emph{Science Robotics},
  vol.~4, no.~31, 2019.

\bibitem{Romiti2023}
E.~Romiti, \emph{et~al.}, ``An optimization study on modular reconfigurable
  robots: Finding the task-optimal design,'' in \emph{IEEE Int. Conf. on
  Automation Science and Engineering (CASE)}, 2023, pp. 1--8.

\bibitem{Kuelz2023}
J.~K\"{u}lz, M.~Mayer, and M.~Althoff, ``{{Timor} {Python}}: A toolbox for
  industrial modular robotics,'' in \emph{Proc. of the IEEE/RSJ Int. Conf. on
  Intelligent Robots and Systems (IROS)}, 2023, pp. 424--431.

\bibitem{Leibrandt2023}
K.~Leibrandt, L.~da~Cruz, and C.~Bergeles, ``Designing robots for reachability
  and dexterity: Continuum surgical robots as a pretext application,''
  \emph{Transactions on Robotics}, vol.~39, no.~4, pp. 2989--3007, 2023.

\bibitem{Liu2020}
S.~B. Liu and M.~Althoff, ``Optimizing performance in automation through
  modular robots,'' in \emph{Proc. of the IEEE Int. Conf. on Robotics and
  Automation (ICRA)}, 2020, pp. 4044--4050.

\bibitem{Hoffman2025}
E.~M. Hoffman, \emph{et~al.}, ``Addressing reachability and discrete
  componentselection in robotic manipulator design through kineto-static
  bi-level optimization,'' \emph{IEEE Robotics and Automation Letters},
  vol.~10, no.~3, pp. 2263--2270, 2025.

\bibitem{mathRobotic}
R.~M. Murray, S.~S. Sastry, and L.~Zexiang, \emph{A Mathematical Introduction
  to Robotic Manipulation}, 1st~ed.\hskip 1em plus 0.5em minus 0.4em\relax CRC
  Press, 1994.

\bibitem{graphicsGems}
R.~Goldman, ``Intersection of two lines in three-space,'' in \emph{Graphics
  Gems}, A.~S. Glassner, Ed., 1994, ch.~3.

\bibitem{rtb}
P.~Corke and J.~Haviland, ``Not your grandmother’s toolbox--the robotics
  toolbox reinvented for {Python},'' in \emph{Proc. of the IEEE Int. Conf. on
  Robotics and Automation (ICRA)}, 2021, pp. 11\,357--11\,363.

\bibitem{PandaSpecificIK}
Y.~He and S.~Liu, ``Analytical inverse kinematics for {F}ranka {E}mika {P}anda
  -- a geometrical solver for 7-{DOF} manipulators with unconventional
  design,'' in \emph{Proc. of the IEEE Int. Conf. on Control, Mechatronics and
  Automation (ICCMA)}, 2021, pp. 194--199.

\bibitem{dktpart2}
J.~Haviland and P.~Corke, ``Manipulator differential kinematics: Part 2:
  Acceleration and advanced applications,'' \emph{Robotics \& Automation
  Magazine}, pp. 2--12, 2023.

\end{thebibliography}

\appendix


\label{sec:newDecompositions}
We derive additional analytical solutions for specific manipulators that were not \daniel{denoted} in \cite{elias_canonical} but are solvable using their solutions to the subproblems in~\tabref{tab:subproblems}.
\daniel{If multiple subproblem decompositions exist for a manipulator, we only present one of them---preferably one where major parts in its derivation can be reused in other manipulator classes.}
The choice of the basis and end effector is interchangeable by the use of kinematic inversion as per \Secref{sec:KinematicInversion}.
Hence, each of the following decompositions (e.g., regarding intersecting axes 5 and 6) is representative for problems with the inversed kinematic chain (e.g., intersecting axes 1 and 2).

The desired orientation $\rotmat{0}{EE}$, and the static rotation between the last joint and the end effector $\rotmat{6}{EE}$ are known. $\rotmat{0}{6}$ is known via \Eqref{eq:r06} and defines the joint rotations via 
\Eqref{eq:forwardOrientationKinRefrained}:
\begin{align}
    \rotmat{0}{6} &= \rotmat{0}{EE}\rotmat{6}{EE}^T
    \label{eq:r06}\\
     &\overset{\eqref{eqn:orientation_fk}}{=} \rotmat{0}{1}\rotmat{1}{2}\rotmat{2}{3} \rotmat{3}{4}\rotmat{4}{5}\rotmat{5}{6}
    \,.
    \label{eq:forwardOrientationKinRefrained}
\end{align}
Without loss of generality, we define our base frame $\frame_0$ such that \mbox{${}^0\vp_1 = \vzero$}.
Consequentially, the position kinematics in~\Eqref{eqn:position_fk} can be rewritten as:
\begin{equation}
  \begin{split}
        \pjnt{1}{6} =& \pjnt{0}{EE} - \pjnt{0}{1} - \rotmat{0}{6}\pjnt{6}{EE}\\ =& 
        {}^0\mR_{1}{}^1\vp_2 + {}^0\mR_{2}{}^2\vp_3 + {}^0\mR_{3}{}^3\vp_4  \\ & + {}^0\mR_{4}{}^4\vp_5 + {}^0\mR_{5}{}^5\vp_6
        \,.
    \end{split}
    \label{eq:forwardPositionKinRefrained}
\end{equation}
We simplify either \Eqref{eq:forwardPositionKinRefrained} or \Eqref{eq:forwardOrientationKinRefrained} and apply one of the subproblems introduced in~\tabref{tab:subproblems} (denoted by ``SP(1\dots4)'') to obtain analytical solutions for a subset of the joint angles.
We repeat the process of reformulation and subproblem-application until the solutions to all joint angles are known.
We denote rotations with known joint angles by~$\rotmatconst{j-1}{j}$.

\paragraph{Spherical Wrist Manipulators}
Let the last three axes of a manipulator intersect in a common point such that ${}^4\vp_5$ and ${}^5\vp_6$ are zero, so that~\Eqref{eq:forwardPositionKinRefrained} becomes
\begin{equation}
    {}^1\vp_6 = \rotmat{0}{1} \left( \pjnt{1}{2} + \rotmat{1}{2}\left(\pjnt{2}{3}+ \rotmat{2}{3}\pjnt{3}{4}\right)\right) \, .
    \label{eq:sphericalWrist}
\end{equation}
This decouples the orientation IK with joint angles ($\theta_4, \theta_5, \theta_6$) from the
position IK with joint angles ($\theta_1, \theta_2, \theta_3$).
We obtain $\theta_4, \theta_5, \theta_6$, and hence $\rotmatconst{3}{4}, \rotmatconst{4}{5}, \rotmatconst{5}{6}$ following the derivations in~\cite[Section 4.1]{elias_canonical}, where the authors also propose solutions to $\theta_1, \theta_2, \theta_3$ for manipulators of class 9.

\xhdrNoPeriod{If the second and third axes intersect}, $\pjnt{2}{3}$ can be set to zero by kinematic remodeling.
As the manipulator has a spherical wrist, we know that $\pjnt{4}{5}=\pjnt{5}{6}=\vzero$.
We can rewrite \Eqref{eq:sphericalWrist} and leverage the norm-preserving property of rotations to obtain:
\begin{align}
    &{}^1\vp_6 = \rotmat{0}{1} \left( \pjnt{1}{2} + \rotmat{1}{2}\left(\rotmat{2}{3}\pjnt{3}{4}\right)\right)\\
    &\Leftrightarrow\rotmat{0}{1}^T{}^1\vp_6 - \pjnt{1}{2}= \rotmat{1}{2}\rotmat{2}{3}\pjnt{3}{4}
    \label{eq:23IntersectingRephrased}\\
     &\Rightarrow0=\left\lVert{}^0\mR_1^T{}^1\vp_6-{}^1\vp_2\right\lVert_2 - \left\lVert{}^3\vp_4\right\lVert_2
    \label{eq:23Intersecting1}
    \,.
\end{align}

\begin{table}
\renewcommand{\arraystretch}{1.3} 
\caption{Subset of the Subproblems presented by Elias et al.~\cite{elias_canonical}. }
\label{tab:subproblems}
\centering
    \begin{tabular}{@{}ll@{}}
    \toprule
    Subproblem & Formulation\\
    \midrule
    SP1 & min $\left\lVert \mR(\vh_i, \theta_i)\vx_1 - \vx_2\right\rVert_2$\\
    SP2 & min $\left\lVert \mR(\vh_i, \theta_i)\vx_1 - \mR(\vh_j, \theta_j)\vx_2\right\rVert_2$\\
    SP3 & min $\left|\left\lVert \mR(\vh_i, \theta_i)\vx_1 - \vx_2\right\rVert_2 - d\right|$ \\
    SP4 & min $\left|\vh_i^T\mR(\vh_j, \theta_j)\vx_1 - d\right|$ \\
    \bottomrule
\end{tabular}
\begin{flushleft}
We denote a vector displacement as $\vx_n \in \mathbb{R}^3$, scalars as $d \in \mathbb{R}$, unit vectors of rotation as $\vh_n \in \mathbb{R}^3$ and corresponding angles as $\theta_n$.
The given formulations are independent of any particular manipulator kinematics.
\end{flushleft}
\renewcommand{\arraystretch}{1} 
\vspace{-2mm}
\end{table}

The solutions we obtain for $\theta_1$ via SP3 from \Eqref{eq:23Intersecting1} pose a superset to the solutions for $\theta_1$ in \Eqref{eq:23IntersectingRephrased}.
The extraneous solutions that arise from such simplifications, e.g., by norm-preservation, can always be found and excluded after inserting them into the original equation once we obtain the remaining joint angles (in this case, $\theta_2$ and $\theta_3$).
Reformulating \Eqref{eq:23IntersectingRephrased} and inserting $\theta_1$ yields \Eqref{eq:23Intersecting2}, from which we obtain $\theta_2, \theta_3$ by applying SP2.
\begin{equation}
    0=\left\lVert{}^1\mR_2^T({\rotmatconst{0}{1}}^T {}^1\vp_6-{}^1\vp_2) - {}^2\mR_3{}^3\vp_4\right\lVert_2
    \label{eq:23Intersecting2}
\end{equation}

\xhdrNoPeriod{If the first and second axis are parallel}, we can choose $\vh_1 =\vh_2$.
A unit vector is invariant to rotations about itself.
This results in the following identities, shown by the example of a rotation about the parallel axes $\vh_1,\vh_2$:
\begin{align}
    \begin{split}
            {}^0\mR_{2}\vh_1 &= {}^0\mR_{1}{}^1\mR_{2}\vh_1={}^0\mR_{1}\vh_1=\vh_1
    \end{split}\\
    \vh_1^T{}^0\mR_{2} &= \vh_1^T{}^0\mR_{1}{}^1\mR_{2}=\vh_1^T{}^1\mR_{2}=\vh_1^T
    \,.
\label{eq:projectiveSimplification}
\end{align}
Hence, multiplying \Eqref{eq:sphericalWrist} by $\vh_1$ or $\vh_2$ simplifies it to:
\begin{align}
    0&=\vh_1^T{}^2\mR_3{}^3\vp_4 - \vh_1^T({}^1\vp_6-{}^1\vp_2 - {}^2\vp_3) \, .
    \label{eq:12Parallel1}
\end{align}
We solve \Eqref{eq:12Parallel1} for $\theta_3$ by using SP4 and arrive at \Eqref{eq:12Parallel2} after rephrasing and employing the L2 norm to \Eqref{eq:sphericalWrist}.
Applying SP3  to \Eqref{eq:12Parallel2} yields $\theta_1$.
\begin{equation}
    0=\left\lVert{}^0\mR_1^T{}^1\vp_6-{}^1\vp_2\right\lVert_2 - \left\lVert\rotmatconst{2}{3}{}^3\vp_4+{}^2\vp_3\right\lVert_2
    \label{eq:12Parallel2}
\end{equation}
Rewriting \Eqref{eq:sphericalWrist} yields \Eqref{eq:12Parallel3}, to which we apply SP1 for $\theta_2$:
\begin{equation}
    0={}^1\mR_2({}^2\vp_3+\rotmatconst{2}{3}{}^3\vp_4) - ({\rotmatconst{0}{1}}^T{}^1\vp_6-{}^1\vp_2)
    \,.
    \label{eq:12Parallel3}
\end{equation}

\paragraph{Three Parallel Axes}
In \cite{elias_canonical}, decompositions for this kinematic class are derived when the second, third, and fourth axes are parallel.
We extend this by proposing:

\xhdrNoPeriod{When the first three axes are parallel and axes five and six intersect}, $\vh_1 = \vh_2 = \vh_3$. We can choose \mbox{$\pjnt{5}{6} = \vzero$} and rewrite \Eqref{eq:forwardPositionKinRefrained} as \Eqref{eq:123Parallel0}.
Multiplying \Eqref{eq:123Parallel0} by $\vh_1$ yields \Eqref{eq:123Parallel1}.
\begin{align}
    \pjnt{1}{6} &= {}^0\mR_{1}{}^1\vp_2 + {}^0\mR_{2}{}^2\vp_3 + {}^0\mR_{3}{}^3\vp_4 + \rotmat{0}{3}\rotmat{3}{4}{}^4\vp_5
    \label{eq:123Parallel0} \\
    0&=\vh_1^T{}^3\mR_4{}^4\vp_5 - \vh_1^T\left({}^1\vp_6 - {}^1\vp_2-\pjnt{2}{3}-\pjnt{3}{4}\right)
    \label{eq:123Parallel1}
\end{align}
We use SP4 on \Eqref{eq:123Parallel1} to obtain $\theta_4$. 
Multiplication of \Eqref{eq:forwardOrientationKinRefrained} by $\vh_5$ and $\vh_1$ leads to \Eqref{eq:123Parallel2}, to which we apply SP4 for $\theta_6$.
Finally, we combine the rotations about the first three axes in a common formulation of SP4 in \Eqref{eq:123Parallel3} for $\theta_{03}=\theta_{1}+\theta_{2}+\theta_{3}$:
\begin{align}
    0&=\vh_5^T{}^5\mR_6{}^0\mR_6^T\vh_1 - \vh_5^T{\rotmatconst{3}{4}}^T\vh_1
    \label{eq:123Parallel2}\\
    0&=\vh_4^T{}^0\mR_3^T{}^0\mR_6{\rotmatconst{5}{6}}^T\vh_5 - \vh_4^T\vh_5
    \,.
    \label{eq:123Parallel3}
\end{align}
We define a unit vector $\vh_n \in \mathbb{R}^3$ that is normal to $\vh_5$, as well as the auxiliary variable $\vdelta \in \mathbb{R}^3$: 
\begin{equation}
  \vdelta = {}^3\vp_4+\rotmatconst{3}{4}{}^4\vp_5 \,.
\end{equation}
Multiplying \Eqref{eq:forwardOrientationKinRefrained} with $\vh_n$ yields  \Eqref{eq:123Parallel4}, from which we obtain $\theta_5$ via SP1.
We use L2 norm-preservation on \Eqref{eq:forwardPositionKinRefrained} to arrive at \Eqref{eq:123Parallel5}, on which we use SP3 for $\theta_2$:
\begin{align}
    0&={}^4\mR_5\rotmatconst{5}{6}\vh_n - {\rotmatconst{3}{4}}^T{\rotmatconst{0}{3}}^T{}^0\mR_6\vh_n
    \label{eq:123Parallel4}\\
    0&=\left\lVert{}^1\mR_2{}^2\vp_3 + {}^1\vp_2\right\lVert_2 - \left\lVert {}^1\vp_6 -\rotmatconst{0}{3}\vdelta\right\lVert_2
    \,.
    \label{eq:123Parallel5}
\end{align}
Using norm-preservation on \Eqref{eq:forwardPositionKinRefrained} yields  \Eqref{eq:123Parallel6}, to which we apply SP1 for $\theta_1$.
We multiply \Eqref{eq:forwardOrientationKinRefrained} by the vector normal to $\vh_3$, $\vh_{n}'$, to arrive at \Eqref{eq:123Parallel7}.
Applying SP1 to \Eqref{eq:123Parallel7} yields $\theta_3$.
\begin{align}
    0&=\left\lVert{}^0\mR_1\left({}^1\vp_2 + \rotmatconst{1}{2}{}^2\vp_3\right) + \rotmatconst{0}{3}\vdelta - {}^1\vp_6\right\lVert_2
    \label{eq:123Parallel6}\\
    0&={}^2\mR_3^T\vh_n' - {\rotmatconst{0}{3}}^T\rotmatconst{0}{1}\rotmatconst{1}{2}\vh_n'
    \,.
    \label{eq:123Parallel7}
\end{align}

\includepdf[pages=-]{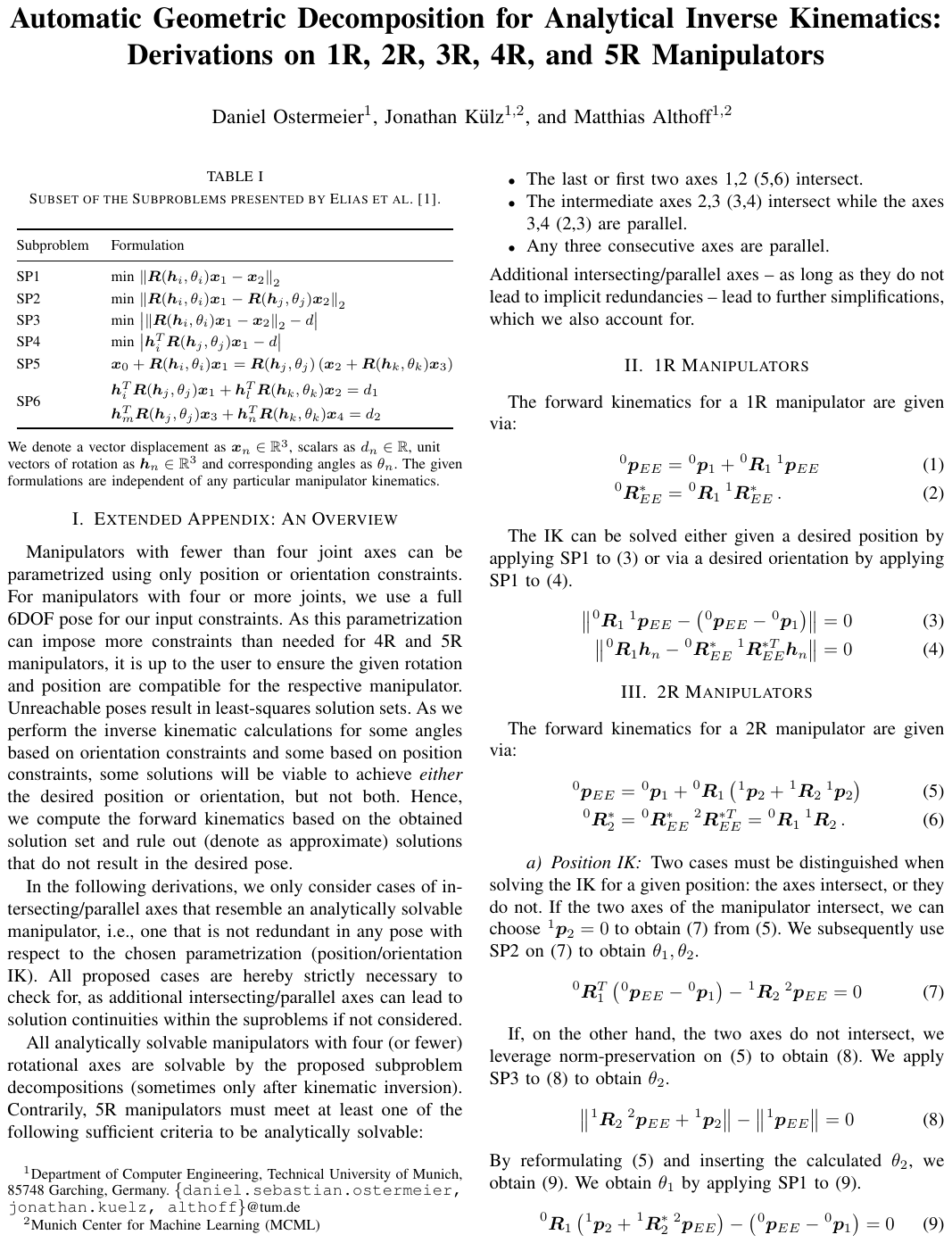}

\end{document}